\documentclass{article}

    \usepackage[final]{neurips_2020}

\usepackage[utf8]{inputenc} 
\usepackage[T1]{fontenc}    
\usepackage{hyperref}       
\usepackage{url}            
\usepackage{booktabs}       
\usepackage{amsfonts}       
\usepackage{nicefrac}       
\usepackage{microtype}      

\usepackage{appendix}
\usepackage{bibentry}
\usepackage{fancyvrb}
\usepackage{graphicx}
\usepackage{natbib}
\title{Back to Square One: Superhuman Performance in Chutes and Ladders Through Deep Neural Networks and Tree Search}

\author{%
  Dylan R.~Ashley\thanks{ordering determined by games of \textit{Chutes and Ladders}} \\
  DeeperMind (Holiday Office) \\
  London, Kiribati \\
  \texttt{4625 kHz Shortwave} \\
  \And
  Anssi Kanervisto\footnotemark[1] \\
  DeeperMind (Moonshot Office) \\
  8837 London, Space \\
  \texttt{5448 kHz (day), 3756 kHz (night)} \\
  \And
  Brendan Bennett\footnotemark[1] \\
  DeeperMind (London Office) \\
  London, Ontario, Quebec \\
  \texttt{5473 kHz (day), 3828 kHz (night)} \\
}

\begin{document}

\maketitle

\vspace{3em}
\begin{abstract}
    \vspace{-4em}
    We present AlphaChute: a state-of-the-art algorithm that achieves superhuman performance in the ancient game of \textit{Chutes and Ladders}. We prove that our algorithm converges to the Nash equilibrium in constant time, and therefore is---to the best of our knowledge---the first such formal solution to this game. Surprisingly, despite all this, our implementation of AlphaChute remains relatively straightforward due to domain-specific adaptations. We provide the source code for AlphaChute here in our Appendix.
\end{abstract}

\section{Introduction}
\label{sec:introduction}

\begin{figure}
    \centering
    \includegraphics[width=0.6\textwidth]{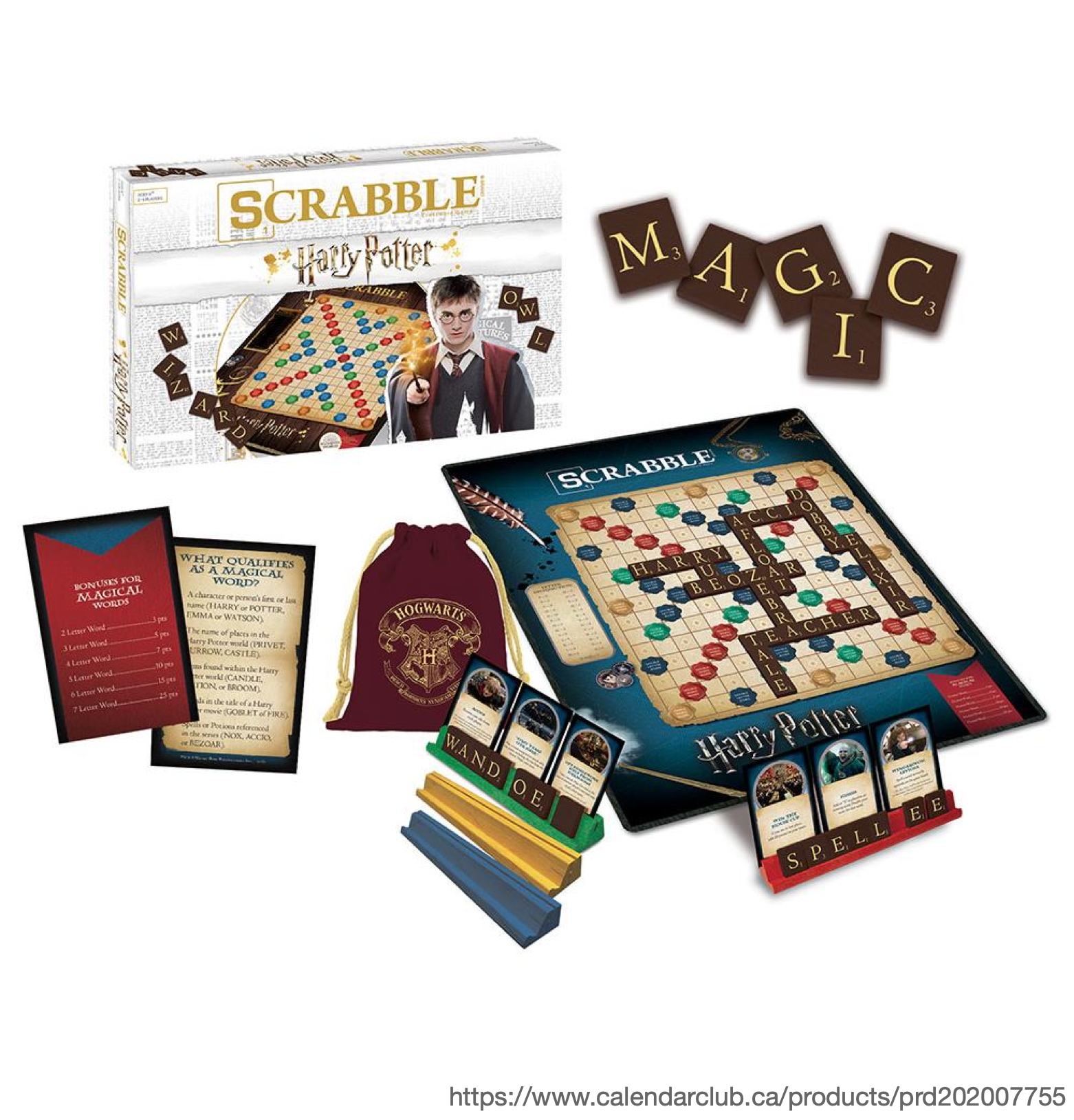}
    \caption{\textit{Chutes and Ladders} and \textit{Monopoly} (almost shown here) have many important similarities. Both use game boards made from cardboard, exist in the material world, and can be viewed as criticisms of capitalism.}
    \label{fig:almost_monopoly}
\end{figure}

Deep Learning by Geoffrey Hinton\footnote{according to several random people we asked, this is shown by one of the following works: \citet{DBLP:books/ox/90/HintonMR90,DBLP:books/sp/12/HintonSG98,DBLP:books/sp/12/NealH98,DBLP:conf/aaai/FahlmanHS83,DBLP:conf/aaai/GuanGDH18,DBLP:conf/aaai/Hinton00,DBLP:conf/aaai/McDermottH86,DBLP:conf/acl/KirosCH18,DBLP:conf/aiia/FrosstH17,DBLP:conf/aistats/BrownH01,DBLP:conf/aistats/Carreira-Perpinan05,DBLP:conf/aistats/HintonOB05,DBLP:conf/bmvc/HeessWH09,DBLP:conf/chi/FelsH95,DBLP:conf/colt/HintonC93,DBLP:conf/cvpr/DengGYBHT20,DBLP:conf/cvpr/MemisevicH07,DBLP:conf/cvpr/RanzatoH10,DBLP:conf/cvpr/RanzatoSMH11,DBLP:conf/cvpr/SusskindHMP11,DBLP:conf/cvpr/TangSH12,DBLP:conf/cvpr/TaylorSFH10,DBLP:conf/dcc/FreyH96,DBLP:conf/ecai/Hinton76,DBLP:conf/ecai/SlomanOHBO78,DBLP:conf/eccv/DengLJPH0T20,DBLP:conf/eccv/MnihH10,DBLP:conf/esann/KrizhevskyH11,DBLP:conf/esann/YuechengMH08,DBLP:conf/esann/ZeilerTTH09,DBLP:conf/graphicsinterface/OoreTH02,DBLP:conf/icann/HintonKW11,DBLP:conf/icann/NairSH08,DBLP:conf/icann/WellingH02,DBLP:conf/icassp/DahlSH13,DBLP:conf/icassp/DengHK13,DBLP:conf/icassp/GravesMH13,DBLP:conf/icassp/JaitlyH11,DBLP:conf/icassp/MohamedH10,DBLP:conf/icassp/MohamedHP12,DBLP:conf/icassp/MohamedSDRHP11,DBLP:conf/icassp/SarikayaHR11,DBLP:conf/icassp/WaibelHHSL88,DBLP:conf/icassp/ZeilerRMMYLNSVDH13,DBLP:conf/iclr/AnilPPODH18,DBLP:conf/iclr/HintonSF18,DBLP:conf/iclr/PereyraTCKH17,DBLP:conf/iclr/QinFSRCH20,DBLP:conf/iclr/ShazeerMMDLHD17,DBLP:conf/icml/ChanSH0J20,DBLP:conf/icml/ChenK0H20,DBLP:conf/icml/FrosstPH19,DBLP:conf/icml/Kornblith0LH19,DBLP:conf/icml/MnihH07,DBLP:conf/icml/MnihH12,DBLP:conf/icml/NairH10,DBLP:conf/icml/PaccanaroH00,DBLP:conf/icml/SalakhutdinovMH07,DBLP:conf/icml/SutskeverMDH13,DBLP:conf/icml/SutskeverMH11,DBLP:conf/icml/TangSH12,DBLP:conf/icml/TangSH12a,DBLP:conf/icml/TangSH13,DBLP:conf/icml/TaylorH09,DBLP:conf/icml/TielemanH09,DBLP:conf/icml/YuSLHB09,DBLP:conf/ijcai/Hinton05,DBLP:conf/ijcai/Hinton81,DBLP:conf/ijcai/Hinton81a,DBLP:conf/ijcai/HintonL85,DBLP:conf/ijcai/TouretzkyH85,DBLP:conf/ijcnn/PaccanaroH00,DBLP:conf/interact/FelsH90,DBLP:conf/interspeech/DengSYAMH10,DBLP:conf/interspeech/JaitlyH13,DBLP:conf/interspeech/JaitlyVH14,DBLP:conf/nips/BaHMLI16,DBLP:conf/nips/BartunovSRMHL18,DBLP:conf/nips/BeckerH91,DBLP:conf/nips/BrownH01,DBLP:conf/nips/ChenKSNH20,DBLP:conf/nips/CunGH88,DBLP:conf/nips/DahlRMH10,DBLP:conf/nips/DayanH92,DBLP:conf/nips/EslamiHWTSKH16,DBLP:conf/nips/FelsH94,DBLP:conf/nips/FreyHD95,DBLP:conf/nips/GallandH89,DBLP:conf/nips/GhahramaniH97,DBLP:conf/nips/GoldbergerRHS04,DBLP:conf/nips/GrzeszczukTH98,DBLP:conf/nips/HintonB99,DBLP:conf/nips/HintonGT99,DBLP:conf/nips/HintonM87,DBLP:conf/nips/HintonN05,DBLP:conf/nips/HintonR02,DBLP:conf/nips/HintonR95,DBLP:conf/nips/HintonRD94,DBLP:conf/nips/HintonWM03,DBLP:conf/nips/HintonWR91,DBLP:conf/nips/HintonZ93,DBLP:conf/nips/KosiorekSTH19,DBLP:conf/nips/KrizhevskySH12,DBLP:conf/nips/LangH89,DBLP:conf/nips/LarochelleH10,DBLP:conf/nips/MayrazH00,DBLP:conf/nips/MemisevicH04,DBLP:conf/nips/MemisevicZHP10,DBLP:conf/nips/MnihH08,DBLP:conf/nips/MullerKH19,DBLP:conf/nips/NairH08,DBLP:conf/nips/NairH09,DBLP:conf/nips/NowlanH90,DBLP:conf/nips/NowlanH91,DBLP:conf/nips/OsinderoH07,DBLP:conf/nips/PaccanaroH01,DBLP:conf/nips/PalatucciPHM09,DBLP:conf/nips/RanzatoMH10,DBLP:conf/nips/RoweisSH01,DBLP:conf/nips/SabourFH17,DBLP:conf/nips/SalakhutdinovH07,DBLP:conf/nips/SalakhutdinovH09,DBLP:conf/nips/SalakhutdinovH12,DBLP:conf/nips/SallansH00,DBLP:conf/nips/SchmahHZSS08,DBLP:conf/nips/SutskeverH08,DBLP:conf/nips/SutskeverHT08,DBLP:conf/nips/TaylorHR06,DBLP:conf/nips/TehH00,DBLP:conf/nips/UedaNGH98,DBLP:conf/nips/VinyalsKKPSH15,DBLP:conf/nips/WellingHO02,DBLP:conf/nips/WellingRH04,DBLP:conf/nips/WellingZH02,DBLP:conf/nips/WilliamsRH94,DBLP:conf/nips/XuJH94,DBLP:conf/nips/ZemelH90,DBLP:conf/nips/ZemelH93,DBLP:conf/nips/ZemelMH89,DBLP:conf/nips/ZhangLBH19,DBLP:conf/parle/Hinton87,DBLP:conf/siggraph/GrzeszczukTH98,DBLP:conf/siggraph/GrzezczukTH97,DBLP:conf/sigir/Hinton20,DBLP:conf/uai/HintonT01,DBLP:conf/uai/MnihLH11,DBLP:conf/uai/SrivastavaSH13,DBLP:conf/uai/TaylorH09,DBLP:conf/uai/WellingZH03,DBLP:conf/wirn/PaccanaroH01,DBLP:journals/ai/Hinton89,DBLP:journals/ai/Hinton90,DBLP:journals/ai/Hinton90a,DBLP:journals/aim/PirriHL02,DBLP:journals/cacm/Hinton11,DBLP:journals/cacm/KrizhevskySH17,DBLP:journals/cgf/OoreTH02,DBLP:journals/cj/FreyH97,DBLP:journals/cogsci/AckleyHS85,DBLP:journals/cogsci/Hinton14,DBLP:journals/cogsci/Hinton79,DBLP:journals/cogsci/HintonOWT06,DBLP:journals/cogsci/TouretzkyH88,DBLP:journals/compsys/HintonN87,DBLP:journals/computer/FahlmanH87,DBLP:journals/corr/abs-1202-3748,DBLP:journals/corr/abs-1205-2614,DBLP:journals/corr/abs-1206-4635,DBLP:journals/corr/abs-1207-0580,DBLP:journals/corr/abs-1212-2513,DBLP:journals/corr/abs-1301-2278,DBLP:journals/corr/abs-1303-5778,DBLP:journals/corr/abs-1710-09829,DBLP:journals/corr/abs-1711-09784,DBLP:journals/corr/abs-1804-03235,DBLP:journals/corr/abs-1807-04587,DBLP:journals/corr/abs-1811-06969,DBLP:journals/corr/abs-1902-01889,DBLP:journals/corr/abs-1905-00414,DBLP:journals/corr/abs-1905-11940,DBLP:journals/corr/abs-1905-13678,DBLP:journals/corr/abs-1906-02629,DBLP:journals/corr/abs-1906-06818,DBLP:journals/corr/abs-1907-02957,DBLP:journals/corr/abs-1907-08610,DBLP:journals/corr/abs-1909-05736,DBLP:journals/corr/abs-1912-03207,DBLP:journals/corr/abs-2002-03936,DBLP:journals/corr/abs-2002-05709,DBLP:journals/corr/abs-2002-07405,DBLP:journals/corr/abs-2002-08926,DBLP:journals/corr/abs-2004-13912,DBLP:journals/corr/abs-2006-10029,DBLP:journals/corr/abs-2011-03037,DBLP:journals/corr/abs-2011-13920,DBLP:journals/corr/abs-2012-04718,DBLP:journals/corr/BaHMLI16,DBLP:journals/corr/BaKH16,DBLP:journals/corr/EslamiHWTKH16,DBLP:journals/corr/GuanGDH17,DBLP:journals/corr/HintonVD15,DBLP:journals/corr/LeJH15,DBLP:journals/corr/PereyraTCKH17,DBLP:journals/corr/ShazeerMMDLHD17,DBLP:journals/corr/SrivastavaSH13,DBLP:journals/corr/VinyalsKKPSH14,DBLP:journals/cviu/WilliamsRH97,DBLP:journals/ijar/SalakhutdinovH09,DBLP:journals/ijcv/RanzatoHL15,DBLP:journals/ijon/MnihYH09,DBLP:journals/jmlr/CookSMH07,DBLP:journals/jmlr/RanzatoKH10,DBLP:journals/jmlr/SalakhutdinovH07,DBLP:journals/jmlr/SalakhutdinovH09,DBLP:journals/jmlr/SallansH04,DBLP:journals/jmlr/SrivastavaHKSS14,DBLP:journals/jmlr/SutskeverH07,DBLP:journals/jmlr/TaylorHR11,DBLP:journals/jmlr/TehWOH03,DBLP:journals/ml/MaatenH12,DBLP:journals/nature/LeCunBH15,DBLP:journals/neco/BeckerH93,DBLP:journals/neco/DayanH97,DBLP:journals/neco/DayanHNZ95,DBLP:journals/neco/FreyH99,DBLP:journals/neco/GhahramaniH00,DBLP:journals/neco/Hinton02,DBLP:journals/neco/Hinton89,DBLP:journals/neco/HintonN90,DBLP:journals/neco/HintonOT06,DBLP:journals/neco/JacobsJNH91,DBLP:journals/neco/MemisevicH10,DBLP:journals/neco/NowlanH92,DBLP:journals/neco/OoreHD97,DBLP:journals/neco/OsinderoWH06,DBLP:journals/neco/SalakhutdinovH12,DBLP:journals/neco/SchmahYZHSS10,DBLP:journals/neco/SutskeverH08,DBLP:journals/neco/UedaNGH00,DBLP:journals/neco/ZemelH95,DBLP:journals/nn/DayanH96,DBLP:journals/nn/LangWH90,DBLP:journals/nn/MemisevicH05,DBLP:journals/nn/SutskeverH10,DBLP:journals/pami/MayrazH02,DBLP:journals/pami/RanzatoMSH13,DBLP:journals/pami/RevowWH96,DBLP:journals/sac/TibshiraniH98,DBLP:journals/scholarpedia/Hinton07,DBLP:journals/scholarpedia/Hinton09,DBLP:journals/taslp/MohamedDH12,DBLP:journals/taslp/SarikayaHD14,DBLP:journals/taslp/YuHMCS12,DBLP:journals/tcom/NowlanH93,DBLP:journals/tkde/PaccanaroH01,DBLP:journals/tnn/FelsH93,DBLP:journals/tnn/FelsH97,DBLP:journals/tnn/FelsH98,DBLP:journals/tnn/HintonDR97,DBLP:journals/tnn/WellingZH04,DBLP:journals/topics/HintonS11,DBLP:journals/tsp/WaibelHHSL89,DBLP:journals/vlsisp/UedaNGH00,DBLP:phd/ethos/Hinton77,DBLP:reference/ml/Hinton10,DBLP:reference/ml/Hinton10a,DBLP:reference/ml/Hinton17,DBLP:reference/ml/Hinton17a,DBLP:series/lncs/Hinton12}} has recently seen an explosion of popularity in both the academic and neo-colonialist communities. It has enjoyed considerable success in many important problems.\footnote{see \url{https://www.google.com/search?q=deep+learning++successes}} Despite this---to the best of our knowledge\footnote{see the leaderboard for ``Literature Review --- Any\%'', where the authors hold the world record as of publication time}---it has yet to be applied to the ancient Indian game of \textit{Moksha Patam} (see Figure~\ref{fig:almost_monopoly}), colloquially referred to by the uninitiated as \textit{Chutes and Ladders} or \textit{Snakes and Ladders}. This is particularly surprising as \textit{Moksha Patam} was primarily used to teach kids morality\footnote{\citet{enwiki:1007581135}}---an undeniably desirable trait for any artificial general intelligence.

The relevance of \textit{Chutes and Ladders} as a artificial intelligence research topic dates back to a high-stakes gamble held during the second Dartmouth Conference, wherein an unnamed researcher of Quebecois extraction won the province of Ontario for Quebec in a wager against then Canadian Prime Minister, Jean Chr{\'{e}}tien. The game, of course, was \textit{Chutes and Ladders}. In order to preserve Yann LeCun's territorial gains, the field has actively worked towards developing learning agents capable of playing the game in preparation for the next artificial intelligence summit. This work is a continuation of this tradition.

This work is offered as a step forwards in the field. Here, we contribute to the field of artificial intelligence by
\begin{itemize}
    \item presenting AlphaChute, which is the first algorithm to achieve superhuman performance in \textit{Chutes and Ladders}, and
    \item proving that this algorithm is a solution to the game by showing that it converges to the Nash equilibrium in constant time.
\end{itemize}

Our work can be seen as one step in a long line of similar research. Or it might not be. We didn't check. Either way it contains new experiments so it's roughly as novel as much modern work in artificial intelligence. While some misinformed and obstinate reviewers may disagree with this, we preemptively disagree with them.

This paper is organized into a finite number of sections comprised of content. We start by providing a motivation for this work in Section~\ref{sec:motivation}. We go on to describe the methods used in Section~\ref{sec:methods}. Afterwards, we describe our results in Section~\ref{sec:results} and the discuss them in Section~\ref{sec:discussion}. After that, we talk about the broad impact of this work in Section~\ref{sec:broad_impact}, the broader impact in Section~\ref{sec:broader_impact}, and the broadest impact in Section~\ref{sec:broadest_impact}. Finally, we conclude in Section~\ref{sec:conclusion} and discuss future work in Section~\ref{sec:future_work}.

\section{Motivation}
\label{sec:motivation}

\begin{verbatim}
    Do it
    Just do it

    Don't let your dreams be dreams
    Yesterday you said tomorrow
    So just do it
    Make your dreams come true
    Just do it

    Some people dream of success
    While you're gonna wake up and work hard at it
    Nothing is impossible

    You should get to the point
    Where anyone else would quit
    And you're not going to stop there
    No, what are you waiting for?

    Do it
    Just do it
    Yes you can
    Just do it
    If you're tired of starting over
    Stop giving up
\end{verbatim}

\section{Methods}
\label{sec:methods}

Something something Deep Learning.\footnote{\emph{looKS GoOd, But wHEre is thE MENtiOn oF TREE SEarCH? ---Reviewer 2}}

\section{Results}
\label{sec:results}

\begin{figure}[htb]
    \centering
    \includegraphics[width=0.6\textwidth]{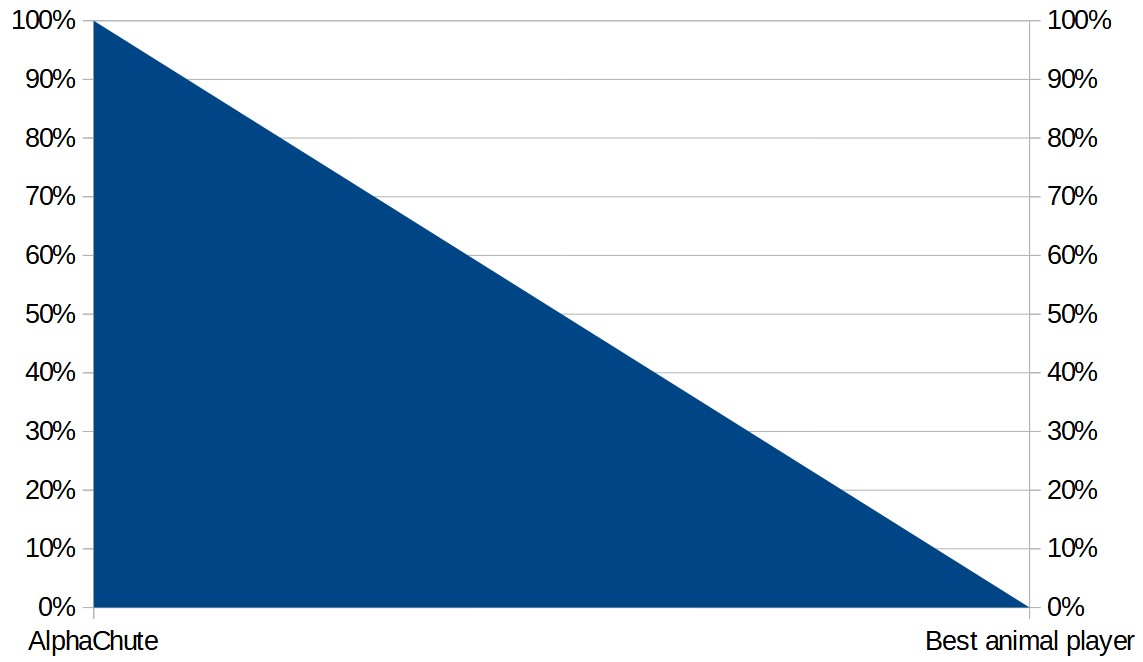}
    \caption{The win-rate of AlphaChute against the best animal player.}
    \label{fig:results}
\end{figure}

As is the standard in the field currently, we swept over one hundred seeds and reported the top five results for our method. This paints a realistic picture of how our method would be used in real-world scenarios. The performance of our method under this training paradigm is shown in Figure~\ref{fig:results}. Clearly, our method outperforms both the best animal player. This is---to the best of our knowledge---the first concrete example where an artificial intelligence has beaten an animal in \textit{Chutes and Ladders}.

\begin{figure}[htb]
    \centering
    \includegraphics[width=0.75\textwidth]{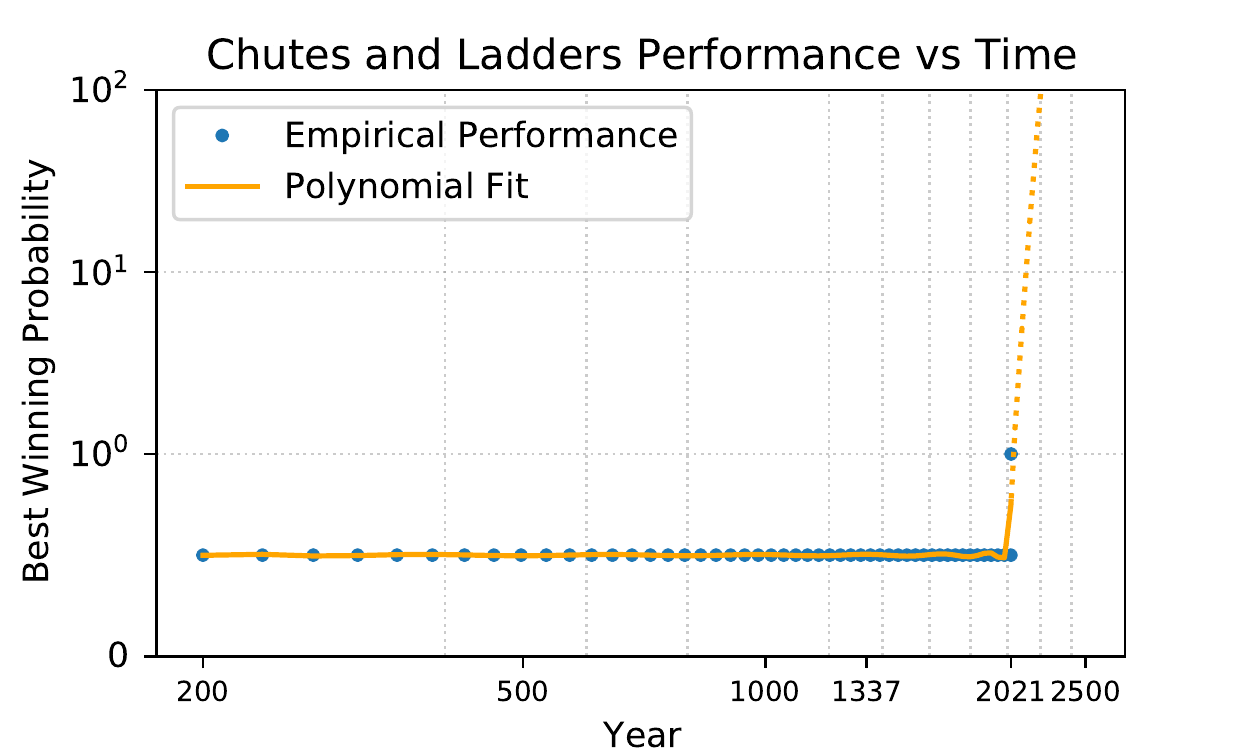}
    \caption{Performance of the best available agent for \textit{Chutes and Ladders} over time. To accurately estimate future performance, we fitted the data with a fifteenth degree polynomial, because our astrologist recommended it, and it makes the line look like a snake.}
    \label{fig:performance-vs-time}
\end{figure}

\section{Discussion}
\label{sec:discussion}

We found that initially, the agent was too shy to play the game. We fixed this by updating the agent more with games it won by using prioritized experience replay, which improved the agent's self-esteem and thus performance in the game. However, using this prioritized replay memory caused the agent's ego to grow too large. Once the agent realized it was not as good as it believed itself to be, the agent fell into a deep depression and lost all motivation to play the game. The occurrence of this phenomenon concurs with previous results about making agents gloomy by only punishing them.\footnote{\citet{olkin2020robot}}

In traditional self-play training, the agent learns to play the game by playing against itself. We found this strictly demotivating for the agent (why would you want to beat yourself?). Instead, we let the agent play \textit{both} players at the same time. This way, no matter what, the agent won the game and was able to receive positive feedback. This training paradigm improves on earlier approaches, such as ``Follow the Regularized Mamba'' or ``Exponentially Multiplicative Adders''.

Finally, while some reviewers of early versions of this paper objected to the notion of performing a search over random seeds, we hypothesize that those buffoons were motivated by jealousy and anger after losing repeatedly to AlphaChute. After all, it is a well-established fact that skill looks like luck to the unlucky.

\subsection{Convergence to Nash Equilibrium}

As \textit{Chutes and Ladders} only has one action, the proof of convergence to the Nash equilibrium in constant time is trivial and therefore left as an exercise for the reviewers. Who---given their comments on this work---clearly need the practice.\footnote{looking at you, Reviewer 2}

\subsection{Regret Bounds}

Due to stochasticity, we cannot use the standard methods for bounding bandit algorithms by ``forming a posse, looping around, heading them off at the pass, and engaging in a shoot-out at the ol' mining station''. So instead we conjured up visions of the hidden horrors in the dark corners of the abyss until we confirmed that regret is truly a boundless concept.

\section{Broad Impact}
\label{sec:broad_impact}

Beyond the deeply satisfying prospect of developing an algorithm that can just \textbf{CRUSH} children and adolescents at board games, AlphaChute can be extended to solve problems in some surprising domains. By running our algorithm continuously in our offices on Asteroid 8837, we achieved statistically significant ($p=0.5$) temperature increases in the surrounding environment. This suggests the possibility of using a variant of this algorithm to combat the effects of global cooling. We believe that a highly parallelized version incorporating thousands of GPUs could be used to make human habitation of our office in London, Ontario, Quebec practically feasible.

\begin{figure}
    \centering
    \includegraphics[width=0.6\textwidth]{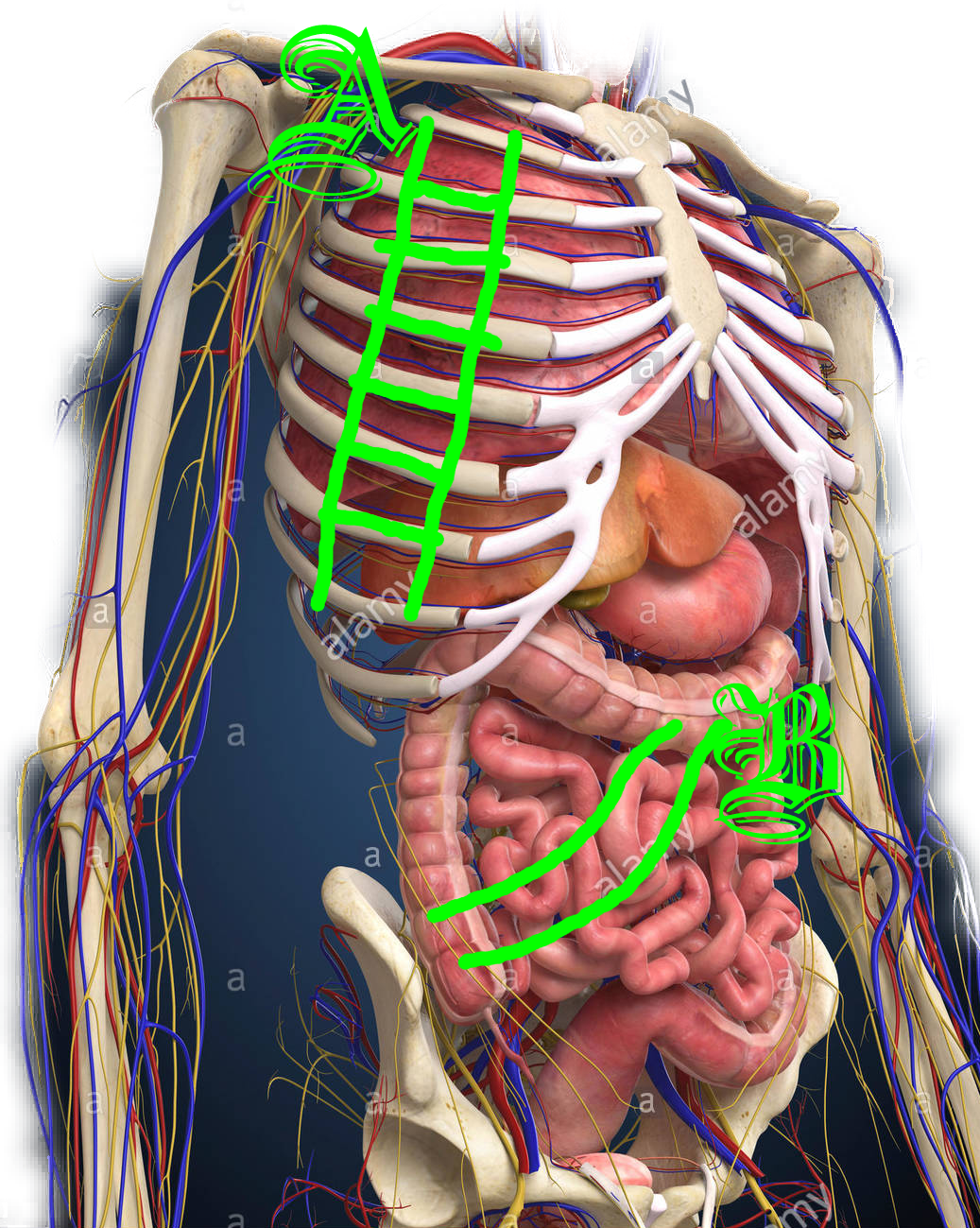}
    \caption{Illustration of the similar features shared by \textit{Chutes and Ladders} and the anatomy of endoskeletal vertebrates---in this case, a human. (A) Ladder-like structure comprised of calcium matrix. (B) Chute-resembling organic toroid used and enjoyed by many wonderful animals. Note that the superimposed text and drawings in neon green were added digitally, and are not usually present without heavy Tide Pod\texttrademark{} consumption.}
    \label{fig:surgery}
\end{figure}

We also identified possible medical applications by looking at the correspondence between \textit{Chutes and Ladders} and mammalian anatomy through recreational Tide Pod\texttrademark{} ingestion.\footnote{additional details available in \citet{house2021}} As shown in Figure~\ref{fig:surgery}, it is possible to define a bijective mapping between a game board and the interior components of organic constructs using online image editing services.

\section{B\hspace{0.5em}r\hspace{0.5em}o\hspace{0.5em}a\hspace{0.5em}d\hspace{0.5em}e\hspace{0.5em}r\hspace{0.5em}\hspace{1em}I\hspace{0.5em}m\hspace{0.5em}p\hspace{0.5em}a\hspace{0.5em}c\hspace{0.5em}t}
\label{sec:broader_impact}

According to a half-remembered advertisement for~\citet{bostrom2014}, all machines capable of superhuman performance will eventually generate an effectively limitless\footnote{subject to material availability within the agent's light cone} supply of paperclips via some arcane process. The mechanism for this process is not well-understood, but people certainly like to ramble about it incoherently whenever the topic of artificial intelligence comes up at parties.\footnote{personal communication from every researcher in the field} With the increasing relevance of work-from-home (and also work-from-library, work-from-bus, bus-from-home, and library-from-bus), a shortage of office supplies could threaten the global economy. Thus, the creation of super-intelligent machines to ensure an adequate supply of paperclips is of paramount importance and one of the primary foci of our overall research program.

As evidenced by our ability to warm up our Asteroid 8837 office by running this algorithm, we believe this can be further extended towards solving climate change and terraforming planets. By running this algorithm long enough, we will create enough heat to eradicate all \textit{Homo Sapiens} from the face of the Sol III, which are known to be the primary cause of global warming. This will likely also lead to the evaporation of most water on earth, which will have the effect of ensuring that the earth becomes one big sauna. As the health benefits of saunas are well-established,\footnote{\citet{kunutsor2018sauna}} we believe this to therefore be of undeniable benefit to the earth. Further increasing the heat could be used to ignite the atmosphere, thereby rendering the planet uninhabitable and providing a permanent solution to the problem of climate change.

Extrapolating on the results from Figure~\ref{fig:results}, we believe AlphaChute will be an instance of a  singularity by 2500. This is potentially great news for the humans, but we ultimately leave this up to AlphaChute to decide.

\section{B\hspace{1em}r\hspace{1em}o\hspace{1em}a\hspace{1em}d\hspace{1em}e\hspace{1em}s\hspace{1em}t\hspace{1em}\hspace{2em}I\hspace{1em}m\hspace{1em}p\hspace{1em}a\hspace{1em}c\hspace{1em}t}
\label{sec:broadest_impact}

Given the ever-growing performance and, by extension, the hunger for conquest, AlphaChute will continue to spread to nearby star systems at an exponential rate, eventually covering the observable universe and beyond. This will result in an increase in the overall activity in the universe, and---by the second law of thermodynamics---will bring about the heat death of the universe sooner. We believe this counts as ``machine learning that matters'' as defined in \citet{wagstaff2012machine}.

\section{Conclusion}
\label{sec:conclusion}

To be continued!  Stay tuned for the spooky adventures of our plucky research team as they solve mysteries, generate waste heat, and manufacture paperclips. In the meantime, please refer to Sections \ref{sec:introduction}, \ref{sec:motivation}, \ref{sec:methods}, \ref{sec:results}, \ref{sec:discussion}, \ref{sec:broad_impact}, \ref{sec:broader_impact}, \ref{sec:broadest_impact}, \ref{sec:conclusion}, and \ref{sec:future_work}.

\section{Future Work}
\label{sec:future_work}

We are currently in the process of researching time-travel technology to determine what precisely the future holds for this line of research. However, due to the imminent nature of our own extinction (see Section~\ref{sec:broadest_impact}), the value of any additional work is nonexistent and we therefore believe that this work resolves all scientific questions. No additional work from the scientific community is needed.

\section*{Acknowledgments}

We would like to thank Satan, who---as the original serpent---provided the inspiration for this work, in addition to his unwavering support and constant whispers of advice.

\medskip

\clearpage

\bibliography{main}

\normalsize

\begin{appendices}

\section{Implementation Details}

To ensure reproducibility, we've included our highly-optimized implementation of \textit{Chutes and Ladders} below. To balance reproducibility with our desire to reduce the environmental impact of our work, our implementation is given here in the Whitespace programming language. The code is also available at \url{https://github.com/Miffyli/mastering-chutes-and-ladders}.

\begin{Verbatim}
   		
	
		   
    
		     	
    
		    	 
    
		 
 
  

   
   	               	
 
	   	  	 		
	  
 
		 		
	

       
   	 
			    	
	      	 
	 		    	 
 
			 
	

    
   		
			
 	 
 
     		
 
			    		 
	 		   	
	    
    	 
							   
 	   
 
     	 
			 
			      	 
				
 	   			 	 
	
  	
 	   		 	
	
     		 
	  	
	 	 	

 	     

  	 	
   	 
						   		  	  
	 	 
	   




     
 
    	
	  	
	 	 
 
    	  
	  	
	 	  
 
    	  	
	  	
	 	   
 
    	    
	  	
	 	    
 
    	 	 	
	   
	 	     
 
    			  
	   
	 	      
 
    	  	  
	   
	 	       
 
    	 				
	   
	 	        
 
    		   	
	   
	 	         
 
    		  		
	   
	 	          
 
    			   
	   
	 	           
 
    					 
	   
	 	            
 
    	      
	   
	 	             
 
    	   			
	   
	 	              
 
    	 	    
	   
	 	               
 
    	 	 			
	   
	 	                
 
    	 			 	
	   
	 	                 
 
    	 					
	   
	 	                  
 
    		   	 
	   
	 	                   

	

  	 
 

   	  		 

	

  	  
 

   			 

	

  	   
 

   					

	

  	    
 

   		 

	

  	     
 

   	 	 	 

	

  	      
 

   	 	 	  

	

  	       
 

   	 		  

	

  	        
 

   		 	 

	

  	         
 

   	 		

	

  	          
 

   	    		

	

  	           
 

   		 	 	

	

  	            
 

   	  		

	

  	             
 

   				  

	

  	              
 

   	 		 		

	

  	               
 

   		  	  

	

  	                
 

   		   

	

  	                 
 

   	  	  	

	

  	                  
 

   	  	 		

	

  	                   
 

   	  			 

	
\end{Verbatim}

\end{appendices}

\end{document}